\pdfoutput=1

\documentclass[11pt]{article}

  \usepackage[preprint]{acl}

\usepackage{times}
\usepackage{latexsym}
\usepackage{amsmath,amssymb,amsfonts}
\usepackage{algorithmic}
\usepackage{graphicx}
\usepackage{textcomp}
\usepackage{xcolor}
\usepackage{algorithm}
\usepackage{booktabs}
\usepackage{makecell}
\usepackage{multirow}
\usepackage{subcaption}
\usepackage{tabularx}
\usepackage[T1]{fontenc}
\usepackage[utf8]{inputenc}
\usepackage{microtype}
\usepackage{inconsolata}
\usepackage{courier}
\usepackage[most]{tcolorbox}
\usepackage{listings}
\usepackage{caption}

\usepackage{tikz}
\usetikzlibrary{positioning,arrows.meta,fit,backgrounds,calc}

\definecolor{codeblue}{rgb}{0.25,0.5,0.75}
\definecolor{codegreen}{rgb}{0.0,0.5,0.0}
\definecolor{codegray}{rgb}{0.5,0.5,0.5}
\definecolor{lightgray}{rgb}{0.9,0.9,0.9}
\definecolor{highlight}{rgb}{0.8,0.95,0.8}

\lstdefinestyle{mystyle}{
    backgroundcolor=\color{white},
    commentstyle=\color{codegreen},
    keywordstyle=\color{codeblue},
    numberstyle=\tiny\color{codegray},
    stringstyle=\color{codeblue},
    basicstyle=\ttfamily\scriptsize,
    breakatwhitespace=false,
    breaklines=true,
    captionpos=b,
    keepspaces=true,
    numbers=left,
    numbersep=5pt,
    showspaces=false,
    showstringspaces=false,
    showtabs=false,
    tabsize=2
}
\title{Dual-Stream LLM-Guided Reward Design for Reject-Inference-Aware\\
Reinforcement Learning in E-Commerce Payment Fraud Detection}

  \author{%
     \textbf{Bo Qu\textsuperscript{\ensuremath{\dagger}}\thanks{Corresponding Author.}},
      \textbf{Zhurong Wang\textsuperscript{\ensuremath{\dagger}}},
    \textbf{Frank Zahradnik\textsuperscript{\ensuremath{\dagger}}},
     \textbf{Ming Ouyang\textsuperscript{\ensuremath{\dagger}}}
    \\
     \textsuperscript{\ensuremath{\dagger}}eBay,
     \\
\texttt{{boqu@ebay.com}}
  }

\begin{document}
\maketitle

\begin{abstract}
Production fraud-detection models learn from a censored slice of their own traffic:
transactions a legacy system blocked never receive fraud labels, so models train on a biased,
approved-only distribution. We extend LLM-guided reward design---the
framework
of \citet{qu2025llmenhanced}, where an LLM evolves the reward of a two-step
REINFORCE fraud-blocking policy---to attack this \emph{reject-inference} problem inside the
reward. A single LLM designs a \textbf{dual-stream reward}: stream~1 applies the standard
dollar-weighted precision objective to labeled transactions; stream~2 scores
legacy-\emph{blocked}, unlabeled transactions by contrasting the newest gradient-boosted risk
signal against the legacy scores that drove the original block. Rather than prescribe which
signals matter, we expose a \emph{superset} of evidence and let the LLM decide for itself which to
use, guided by evaluation feedback alone; because it writes the reward as code, that choice is
directly auditable.
On a held-out four-month out-of-time test, the dual-stream agent beats the prior
labeled-only agent at every target recall---by $+4.7$, $+2.4$ and $+3.0$ points of
dollar-weighted precision at 80\%, 85\% and 90\% recall---and the production baseline by $+6.8$
points on average. The resulting policy is architecturally identical to
\citet{qu2025llmenhanced} and adds no inference cost.
\end{abstract}

\section{Introduction}

\begin{figure*}[t]
\centering
\definecolor{promptbg}{RGB}{214,229,252}
\definecolor{llmbg}{RGB}{252,238,210}
\definecolor{dualstreambg}{RGB}{210,245,218}
\resizebox{0.88\textwidth}{!}{%
\begin{tikzpicture}[
  >=Latex,
  node distance=0.5cm,
  box/.style={draw, rounded corners=3pt, align=center, inner sep=4pt, font=\footnotesize},
  arr/.style={->, thick},
  darr/.style={->, dashed, semithick},
  annotarr/.style={->, thin, gray!70!black, shorten >=2pt, shorten <=2pt},
]
\node[box, fill=promptbg, text width=1.85cm, minimum height=2.8cm] (prompt) {%
  \textbf{Prompt}\\
  \textbf{Context}\\[2pt]
  \scriptsize task desc.\\
  RL env code\\
  human reward\\
  fraud-rate\\
  statistics%
};
\node[box, fill=llmbg, right=0.6cm of prompt, text width=1.55cm, minimum height=2.8cm] (llm) {%
  \textbf{LLM}\\[2pt]
  \scriptsize Qwen3-class\\
  (open, FP8)\\
  $1{\times}$A100\\
  80\,GB%
};
\node[box, fill=white, right=0.6cm of llm, text width=1.65cm, minimum height=2.8cm] (ast) {%
  \textbf{Candidate}\\[2pt]
  \scriptsize \texttt{get\_reward}\\
  (full sig.)\\[3pt]
  AST check;\\
  resample\\
  if invalid%
};
\node[box, fill=white, right=1.0cm of ast, yshift=1.1cm, text width=2.2cm] (stream1) {%
  \textbf{Stream\,1: Labeled}\\[2pt]
  \scriptsize \texttt{appr\_flag=1},\\
  target $\in\{0,1\}$\\
  Lagrangian prec.\\
  reward (human baseline)%
};
\node[box, fill=white, below=0.45cm of stream1, text width=2.2cm] (stream2) {%
  \textbf{Stream\,2: Blocked}\\[2pt]
  \scriptsize \texttt{appr\_flag=0},\\
  target $=-1$, censored\\
  newest vs.\ legacy GBMs;\\
  \emph{LLM selects signals}%
};
\begin{pgfonlayer}{background}
  \node[draw, dashed, rounded corners=5pt, fill=dualstreambg,
        fit=(stream1)(stream2), inner sep=7pt] (dualbox) {};
\end{pgfonlayer}
\node[font={\footnotesize\bfseries}, anchor=south west]
  at ($(dualbox.north west) + (3pt, 2pt)$) {Dual-Stream Reward};
\node[box, fill=white, right=0.7cm of dualbox, text width=1.95cm, minimum height=2.8cm] (policy) {%
  \textbf{REINFORCE}\\[2pt]
  \scriptsize \textbf{Step\,0 (PRA)}: all\\
  \textbf{Step\,1 (PA)}: passed\\[2pt]
  newest GBMs\\
  $+$ embed.$\to$softmax\\[2pt]
  $-\!\log\pi{\cdot}R$;\\
  ${\sim}200$\,ep./cand.%
};
\node[box, fill=white, right=0.6cm of policy, text width=1.55cm, minimum height=2.8cm] (eval) {%
  \textbf{Eval}\\[2pt]
  \scriptsize in-loop: 1\,mo.\\
  final: 4\,mo.\ OOT\\[2pt]
  \$-wtd.\ prec.\\
  @\,target recall%
};
\draw[arr] (prompt) -- (llm);
\draw[arr] (llm) -- (ast);
\draw[arr] (ast.east) -- (dualbox.west);
\draw[arr] (dualbox.east) -- node[above, font=\scriptsize, fill=white, inner sep=1pt]{shared grad.} (policy.west);
\draw[arr] (policy) -- (eval);
\draw[darr, gray!60!black] (eval.north) -- ++(0,2.2) -| (llm.north);
\node[font=\scriptsize, align=center, text width=3.5cm]
  at ($(eval.north)!0.5!(llm.north) + (0,2.45)$)
  {best\,/\,suboptimal\,/\,reflection\\
   adapt.\ temp.\ (${\times}0.9$\,/\,${\times}1.1$)};
\node[font=\scriptsize, align=center, text width=2.6cm, above=1.3cm of ast]
  {${\sim}40$ iter.\ ${\times}{\sim}10$ cand.;\\best reward retained};
\node[font=\scriptsize, align=center, text width=2.2cm, text=gray!70!black,
      below=0.45cm of stream2] (legannot)
  {\textit{legacy GBM scores:}\\
   \textit{reward-only; never}\\
   \textit{in policy input state}};
\draw[annotarr] (stream2.south) -- (legannot.north);
\node[font=\scriptsize, align=center, text width=1.9cm, text=gray!70!black,
      below=0.45cm of eval]
  {\textit{policy net unchanged}\\(no added cost)};
\end{tikzpicture}%
}
\caption{%
  Dual-Stream LLM-Guided Reward Design. An open-weight Qwen3-class LLM (FP8, single
  A100\,80\,GB) evolves candidate \texttt{get\_reward} functions with adaptive temperature,
  retaining the best. Each AST-validated candidate scores two streams under one REINFORCE
  gradient and one two-step policy: \textbf{Stream\,1}, the dollar-weighted Lagrangian on
  labeled approved rows; \textbf{Stream\,2}, a censored-outcome reward on historically blocked
  rows whose label is unobservable. The prompt exposes a \emph{superset} of candidate signals and
  the LLM selects which to rely on from evaluation feedback (\S\ref{sec:whatlearned}). Legacy
  scores enter \emph{only} the reward, so the trained policy is architecturally identical to the
  labeled-only baseline and adds no inference cost.%
}
\label{fig:LLM-RL-overview}
\end{figure*}
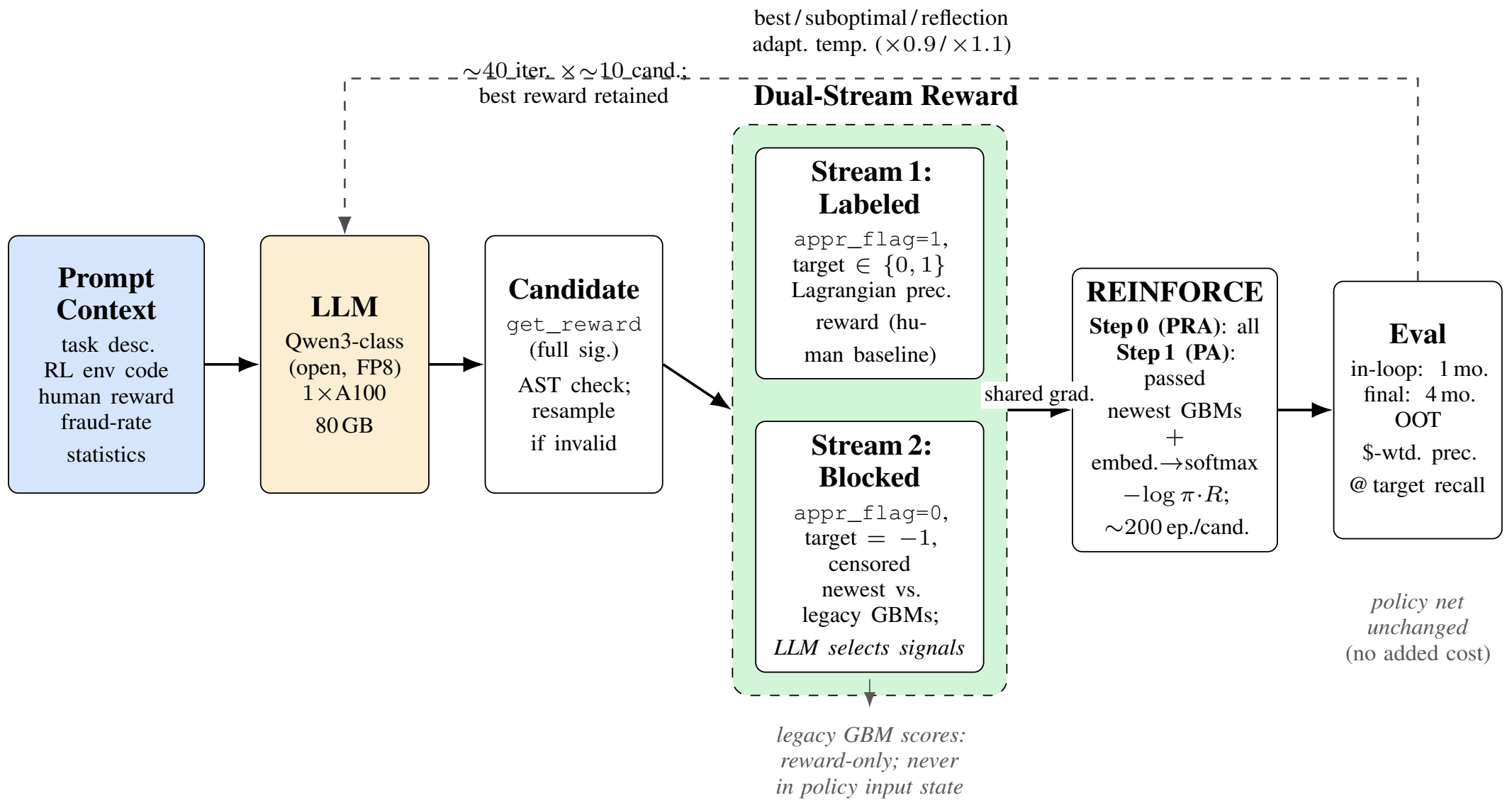

Large language models (LLMs) now match or exceed human performance on code generation
\cite{chen2021evaluating} and reasoning \cite{kojima2022large}, and a growing line of work uses
them to \emph{design the reward functions} of RL agents automatically: prompting an LLM as a proxy
reward evaluator \cite{kwon2023reward}, and the evolutionary code-generation paradigm of EUREKA
\cite{ma2023eureka}, Text2Reward \cite{xie2024text2reward} and Language-to-Rewards
\cite{yu2023language}---the LLM writes reward code, a policy is trained per candidate, and feedback
conditions the next generation---later extended with reward critics \cite{li2024automc},
sim-to-real transfer \cite{ma2024dreureka} and sample-efficient reformulations
\cite{sarukkai2024automated,sun2025card}. This literature is dominated by robotics and game
benchmarks, where every rollout yields a reward.
Prior work brought this paradigm to e-commerce payment fraud: \citet{qu2025llmenhanced} frames transaction risk as a multi-step Markov decision process (MDP)
and lets an LLM evolve the reward for a two-step REINFORCE policy that blocks or passes each
transaction at pre- and post-authorization, outperforming both human-designed Lagrangian
rewards and the production gradient-boosted-machine (GBM) baseline in dollar-weighted
precision at target recall.

\paragraph{The censored-label gap.}
That system, like essentially all supervised fraud models, trains on \emph{approved}
transactions only---those the legacy model and rules let through, whose fraud outcome (a
chargeback) is eventually observed. But a large fraction of traffic is \emph{blocked} before
it can produce an outcome. These blocked rows are scored by the same newest GBM models that
feed the RL policy and carry the legacy scores that drove the block, yet their ground-truth
label is fundamentally \emph{unobservable}. Training on the approved slice alone thus
optimizes on a distribution systematically narrower---and lower-risk---than the deployment
population. This is the classical \emph{reject-inference} / selective-labels problem of
credit scoring \cite{heckman1979sample,lakkaraju2017selective,ehrhardt2021reject}, here
appearing inside an LLM-guided RL reward-design loop for the first time. Classical remedies
attack it with explicit machinery---Heckman selection correction, self-labeling
\cite{kozodoi2020shallow}, reject-aware multi-task networks \cite{liu2022rmt}, and
positive-unlabeled estimators \cite{elkan2008learning,duplessis2015convex,bekker2020learning};
we instead let one LLM reason over each blocked row's newest-versus-legacy score structure and
\emph{synthesize} the censored-outcome reward directly.

\paragraph{Positioning.}
Our reward sits in the cost-sensitive, constrained-RL tradition rather than the data- or
loss-level imbalance remedies \cite{chawla2002smote,lin2017focal}: fraud as sequential decisioning
\cite{shen2020deep,vimal2021application,qayoom2024novel,henderson2023irs}, a per-stage precision
constraint in the spirit of \citet{tessler2019rcpo}, a dollar-weighted objective
\cite{dervovic2023vdr}, and cost-sensitive decision theory
\cite{elkan2001foundations,bahnsen2015edcst,hoppner2022idcs,verbraken2013emp,cotter2019nondiff}.
It inverts RLHF/RLAIF \cite{ouyang2022instructgpt,bai2022constitutional,lee2024rlaif}---the LLM
\emph{designs} a reward for an external policy and is never updated---aligning with code-as-policy
work \cite{liang2023cap,wang2023voyager}. LLMs appear in industrial trust-and-safety pipelines
\cite{qiao2024scaling,chakraborty2024detoxbench,yang2025flag,fan2025enhancing,kath2025llm} and
e-commerce fraud detection \cite{rodrigues2022fraud,mutemi2024ecommerce}; none applies LLM-guided
reward design to censored fraud labels.

\paragraph{Contributions.}
\begin{itemize}
\item \textbf{Reject-inference-aware reward design.} To our knowledge, the first
LLM-guided reward-design loop to learn from censored, never-approved samples. Prior LLM
reward work \cite{ma2023eureka,li2024automc,yu2023language} targets robotics and games,
where every rollout yields a reward; none confronts labels missing by construction.
\item \textbf{A dual-stream reward, designed semi-zero-shot.} One LLM emits a single reward with
two internal streams (Figure~\ref{fig:LLM-RL-overview}): stream~1 keeps the human baseline
Lagrangian on labeled rows, preserving the production objective; stream~2 scores blocked rows from
their newest-versus-legacy score structure. Stream~1 is seeded from the reference reward
\citet{qu2025llmenhanced} refined under its few-shot setting, but no human design exists for the
blocked stream, so stream~2 is authored \emph{zero-shot}---evidence that the loop can
\emph{invent} reward structure, not only tune it.
\item \textbf{Signal selection left to the LLM, and auditable afterwards.} Rather than
hand-encoding which risk signals matter, we expose a superset of 14 candidates and let evaluation
feedback drive the choice. The selected reward relies on two of them and ignores
the rest---including the legacy scores our own schema assumed it would need---and because it is
emitted as code, that decision is inspectable rather than buried in fitted weights (\S\ref{sec:whatlearned}).
\item \textbf{Held-out temporal evidence, month by month.} We evaluate on a four-month
out-of-time window held out from \emph{both} policy training and the LLM selection loop, reporting
each month at each target recall rather than only pooled. At the tightest operating point the
dual-stream reward leads the published method of \citet{qu2025llmenhanced} in all four months; at
loose recall the ordering reverses, which \S\ref{sec:monthly} traces to a difference in \emph{where}
each policy does its blocking.
\item \textbf{A deployment-realistic setup.} The reward designer is an open
$\sim$27B-parameter model (Qwen3-class, FP8) on a single A100 (80\,GB); the resulting policy is
unchanged in size and latency from the work of \citet{qu2025llmenhanced}.
\end{itemize}

\section{Method}

\subsection{Transaction Risk as a Constrained MDP}
This work adopts the MDP formulation of \citet{qu2025llmenhanced}. A transaction passes through two
checkpoints---\emph{pre-authorization} (PRA, step~0), where the agent acts on all
transactions, and \emph{post-authorization} (PA, step~1), where it acts only on
transactions it passed at step~0. At step $i$ the state $\mathcal{S}_i$ is the vector of
newest GBM risk scores plus a stage indicator, the action $\mathcal{A}_i \in
\{\text{pass},\text{block}\}$, and the reward $\mathcal{R}_i = R(\mathcal{S}_i,
\mathcal{A}_i)$. We maximize the business objective
\begin{align}
\label{equ:opt_target}
\text{Maximize } & \; \text{\$TP} - \text{\$FP} \\
\text{subject to } & \; \text{\$TP}_{\text{stage 1}} > \text{\$TP}_{\text{stage 2}}, \nonumber
\end{align}
i.e.\ maximize dollar-weighted true positives net of false positives while blocking more
fraudulent GMV early (PRA) than late (PA). The human baseline reward follows from a
Lagrangian relaxation of a per-stage precision constraint $\text{\$TP}_i / (\text{\$TP}_i
+ \text{\$FP}_i) > \alpha_i$:
\begin{equation}
\label{equ:rw_1}
R_{\text{precision}}^{i}(s,a) = (1-\alpha_i)\,\text{\$TP}_i - \alpha_i\,\text{\$FP}_i .
\end{equation}
The policy is a compact DNN (categorical embeddings $+$ GBM scores $\rightarrow$ two-class
softmax) trained with REINFORCE \cite{williams1992simple} (Adam, lr $10^{-4}$).

\subsection{Censored Outcomes and the Dual-Stream Reward}
\label{sec:dualstream}
In the dual-stream method the training set is the union of two streams:
\begin{itemize}
\item \textbf{Labeled / approved} rows ($\texttt{appr\_flag}=1$), with an observed fraud
label $\texttt{target}\in\{0,1\}$.
\item \textbf{Blocked} rows ($\texttt{appr\_flag}=0$), with $\texttt{target}=-1$
(unobservable). These carry legacy risk scores (two model generations $\times$ four heads)
that informed the original blocking decision, together with buyer segment and payment
method.
\end{itemize}
We denote the four legacy supervised risk-model heads $h_1$--$h_4$ and the legacy model
generations \texttt{gen-1}, \texttt{gen-2}; the two-generation count is incidental---the
dual-stream reward accepts any number of legacy score sets, each an additional
staleness-aware evidence signal on blocked rows.
The LLM designs a single reward with the extended \textbf{19-argument} signature
\begin{quote}\small\ttfamily
get\_reward(current\_step, action, target, wgt, is\_blocked, legacy\_g1\_\{h1,h2,h3,h4\},
legacy\_g2\_\{h1,h2,h3,h4\}, newest\_\{h1,h2,h3,h4\}, byr\_segment, pymnt\_inst)
\end{quote}
On labeled rows ($\texttt{is\_blocked}=0$) the legacy and context arguments are zero, so
stream~1 is a function of $(\text{current\_step}, \text{action}, \text{target},
\text{wgt})$ alone and takes the same dollar-weighted Lagrangian form as
Eq.~\ref{equ:rw_1}, with the LLM free to co-tune its coefficients. Because stream~2 is
gated entirely by $\texttt{is\_blocked}$, it never contributes to the reward on approved
traffic, so the resulting policy remains architecturally identical to the labeled-only
baseline. On blocked rows the LLM designs stream~2 \emph{from scratch}: unlike stream~1, which is
seeded from the human baseline reward of Eq.~\ref{equ:rw_1}---the same reference \citet{qu2025llmenhanced}
refined under its few-shot setting---the reject-inference stream has no human-authored template
to seed from, so its entire structure is produced zero-shot. The dual-stream method must therefore start the
blocked stream in the zero-shot regime rather than refining a human design.

\paragraph{Legacy scores are reward-only.}
A deliberate architectural choice: legacy scores enter the reward function \emph{only}---never
the policy's input state. The policy learns from the newest real-time signal, while the legacy
scores serve as reward-time \emph{evidence} about a blocked row's latent risk, keeping the
policy's feature contract unchanged and preventing it from depending on signals it
would not see at scoring time.

\paragraph{Score disagreement as an uncertainty signal.}
For a blocked row the newest GBM score and the legacy score jointly form a natural
uncertainty proxy: agreement at high risk supports a confident block; a high legacy score
contradicted by a low newest score suggests over-blocking and a conservative reward; a low
legacy score revised upward signals newly-caught risk; and mutual agreement at low risk
drives the reward toward a pass. Because one LLM reasons over both signals semantically, no
separate PU-learning or inverse-propensity estimator is needed---the disagreement structure
is already in the data and the LLM synthesizes it, sidestepping the importance-weighting or
self-labeling machinery that classic reject inference and PU learning employ
\cite{elkan2008learning,kozodoi2020shallow,liu2022rmt}.

\paragraph{A menu of signals, not a claim about them.}
On blocked rows we deliberately expose a \emph{superset} of candidate evidence---both legacy
generations across all four heads, the four newest heads, buyer segment and payment instrument, 14
arguments in all---without claiming in advance that any is informative. Exposing them \emph{is}
the experiment: because the reward is code, the evolved function records which signals the designer
leaned on and which it discarded (\S\ref{sec:whatlearned}).

\paragraph{Step routing and aggregation.}
At step~0 stream~2 uses the PRA legacy and newest scores; at step~1 (only rows passed at
step~0) the PA scores. The two-step rewards are summed and the REINFORCE loss
$-\log\pi \cdot R$ accumulated across both steps, exactly as in stream~1, so the two streams
share one gradient and one policy.

\subsection{LLM Reward Optimization Loop}
The evolutionary loop is unchanged from that of \citet{qu2025llmenhanced} and is summarized in
Algorithm~\ref{alg:LLM-RL} (Appendix~\ref{appendix:alg}): over $N_\text{iter}$ outer iterations it samples $N_\text{samples}$
candidate reward functions per iteration; each valid candidate (checked by AST parsing
against the required signature) trains a fresh policy for $N_\text{episodes}$ episodes; the
best candidate by dollar-weighted precision at target recall is retained and its performance
is fed back into the LLM context for the next iteration. LLM temperature is adaptive (decayed
on improvement, grown on plateau); loop constants are given in \S\ref{sec:setup}. To ground stream~2 we additionally inject a compact table of
\emph{dollar-weighted fraud rates by buyer cohort and payment method}, computed on the labeled
training data, so the LLM can calibrate context-conditioned weights rather than guessing them.
The full prompt templates are reproduced in Appendix~\ref{appendix:prompts}.

\section{Experimental Setup}
\label{sec:setup}

\paragraph{Data.}
All data are real-world transactions drawn from a large e-commerce payment system, split into
three disjoint temporal windows: policy training on February~2026, LLM candidate selection on
March~2026, and a held-out out-of-time (OOT) test on April~1--July~31, 2026. The fraud target is an
unauthorized-chargeback label matured as of 2026-09-01. Training combines ${\sim}2.06$M labeled
approved transactions with ${\sim}3.91$M legacy-blocked ones ($\approx\!1.9\times$ the labeled
stream), for ${\sim}5.97$M samples---nearly tripling the usable data. March drives LLM
best-candidate selection; the four OOT months are held out from both training and selection and
are used only for the evaluation reported below.

\paragraph{Reward-designer LLM and hardware.}
The reward designer is an open Qwen3-class $\sim$27B-parameter instruction model served in
FP8. All training and LLM inference run on a single \textbf{NVIDIA A100 (80\,GB)} with
standard CPU/RAM support---a substantial simplification over the setup of \citet{qu2025llmenhanced} (2$\times$
V100 32\,GB, 4-bit), enabled by FP8 serving of the larger backbone.

\paragraph{Loop hyperparameters.}
Both systems run the loop of \citet{qu2025llmenhanced} under one shared configuration, adjusted
from that work for a direct comparison: $N_\text{iter}=40$,
$N_\text{samples}=10$, $N_\text{episodes}=200$, optimizing around target recalls
$\theta_\text{recall}\in[80\%,85\%,90\%]$, policy training batch $=10{,}240$, Adam
lr $=10^{-4}$, adaptive temperature $0.9/1.1$.
We compare two LLM-designed systems: the labeled-only \emph{single-stream} agent (\textbf{SS}),
reproducing \citet{qu2025llmenhanced}, and the \emph{dual-stream} agent (\textbf{DS}) of this
work. Both ran the \emph{same} full $N_\text{iter}=40$ budget with no early stopping; their
best-candidate iterations (SS~34, DS~25) mark only \emph{where} each run's top reward emerged, so
the comparison is not confounded by run length.

\paragraph{Metric.}
Dollar-weighted \$Precision at fixed dollar-weighted \$Recall thresholds, computed against
the production baseline risk score $\mathcal{M}_b$ (the legacy production score at the PRA
checkpoint, written \textbf{Base} in all tables). The two-step agent's operating points are
obtained by sweeping block-score thresholds across both stages; because each system's sweep lands
on its own discrete grid, the achieved recalls differ slightly between systems at a given target.
The SS and DS agents reported in Section~\ref{sec:results} are each run's best candidate on the
March evaluation set.

\section{Results}
\label{sec:results}

\subsection{Held-Out OOT Generalization}
Table~\ref{tab:oot_overall} reports overall dollar-weighted precision on the four-month held-out
window (April--July~2026) at target recalls 0.80--0.90 (\S\ref{sec:setup}).

Across this range the dual-stream agent improves on the production baseline at every target:
\$precision rises from 47.09\% to \textbf{55.15\%} at 80\% recall, from 39.95\% to
\textbf{46.36\%} at 85\%, and from 29.91\% to \textbf{35.92\%} at 90\%---gains of $+8.06$,
$+6.41$ and $+6.01$ points, $+6.83$ on average. It also beats the labeled-only agent at every
target, by $+4.74$, $+2.36$ and $+2.97$ points ($+3.36$ on average), so DS dominates SS across
the whole reported range rather than winning only on the mean; Appendix~\ref{appendix:oot} gives
the full swept curve for both agents. We conducted three independent training trials for each agent; results stabilized across
trials, with exact operating points varying trial to trial.

\paragraph{Where the fraud is stopped.} Eq.~\ref{equ:opt_target} also requires that more
fraudulent GMV be stopped pre- than post-authorization. This is an \emph{aggregate} constraint,
imposed on the reward in expectation over the training distribution rather than enforced at any
particular threshold pair; nothing guarantees it at every operating point an agent is later swept
to, so the meaningful question is how often, and how badly, a trained policy breaks it. The
right-hand columns of Table~\ref{tab:oot_overall} give the share of each agent's blocked fraud GMV
intercepted at step~0. DS keeps the majority pre-authorization at all three targets (67.1\%,
94.8\%, 64.8\%); SS does so at two of the three, stopping only \textbf{12.5\%} early at the 90\%
target and routing the remainder past the issuer authorization request. On this evaluation window,
then, precision and step ordering do not appear to trade off against one another.

\begin{table}[t]
\small
\centering
\setlength{\tabcolsep}{3pt}
\caption{Overall held-out OOT (April--July~2026), read at the first operating point at or above the
target \$Recall; no interpolation. \textbf{Base}: production GBM. \textbf{SS}: labeled-only agent
reproducing \citet{qu2025llmenhanced}. \textbf{DS}: dual-stream (this work). Right-hand block:
share of each agent's \emph{blocked fraud GMV} intercepted at step~0;
Eq.~\ref{equ:opt_target} asks for $>$50\%; the underlined cell falls below it.}
\label{tab:oot_overall}
\begin{tabular}{lccc@{\hskip 5pt}cc}
\toprule
 & \multicolumn{3}{c}{\$Precision} & \multicolumn{2}{c}{\$TP stopped at PRA} \\
\cmidrule(lr){2-4}\cmidrule(lr){5-6}
\$Recall & Base & SS & DS & SS & DS \\
\midrule
~80\% & 47.09\% & 50.41\% & \textbf{55.15\%} & 64.8\% & \textbf{67.1\%} \\
~85\% & 39.95\% & 44.00\% & \textbf{46.36\%} & 81.6\% & \textbf{94.8\%} \\
~90\% & 29.91\% & 32.95\% & \textbf{35.92\%} & \underline{12.5\%} & \textbf{64.8\%} \\
\midrule
avg & 38.98\% & 42.45\% & \textbf{45.81\%} & & \\
\multicolumn{4}{l}{targets satisfying Eq.~\ref{equ:opt_target}} & 2\,/\,3 & \textbf{3\,/\,3} \\
\bottomrule
\end{tabular}
\end{table}

\subsection{Month-by-Month Behavior}
\label{sec:monthly}
Pooled results can mask a policy that decays as it ages away from its training window.
Table~\ref{tab:oot_month} breaks the held-out window into its four months, each evaluated
independently and reported \emph{at each target recall}---the agents behave differently across the
band, and averaging hides it. Absolute levels shift sharply month to month (at 80\% the baseline
runs 61.27\% in April against 19.27\% in May), so only within-row comparisons are meaningful.

Two patterns hold. First, the dual-stream advantage over SS \emph{concentrates at the tight end of
the band}: at the 80\% target DS leads SS in all four months, while at 90\% the ordering reverses
and SS leads in three of four. Against the production baseline DS is positive every month at 85\%
and 90\% and in three of four at 80\%---where SS clears the baseline only once, falling $5.39$
points below it in July. Second, that reversal is the step-ordering result of \S\ref{sec:results}
seen from another angle: SS earns its loose-recall precision by deferring blocking to
post-authorization, intercepting just 12.5\% of blocked fraud GMV early at the pooled 90\% target
(Table~\ref{tab:oot_overall})---strong on precision alone, but it pushes fraud past the issuer
authorization request. Together the two tables place DS ahead where a payment system would actually
operate: the tight, high-precision end of the band, without giving up early interception.

\begin{table}[t]
\footnotesize
\centering
\setlength{\tabcolsep}{4pt}
\caption{Held-out OOT \$Precision \emph{differences} by month and target \$Recall (R), same
extraction rule as Table~\ref{tab:oot_overall}. Absolute levels are omitted: the fraud mix shifts
month to month, so only within-cell system comparisons are meaningful. Bold marks a DS win.}
\label{tab:oot_month}
\begin{tabular}{lcccc}
\toprule
R & Apr & May & Jun & Jul \\
\midrule
\multicolumn{5}{l}{\textit{DS $-$ SS}}\\
~80\% & \textbf{$+$2.80} & \textbf{$+$1.00} & \textbf{$+$1.91} & \textbf{$+$4.77} \\
~85\% & \textbf{$+$2.83} & $-$5.28 & $-$1.15 & \textbf{$+$4.97} \\
~90\% & $-$2.08 & $-$1.36 & $-$7.23 & \textbf{$+$3.81} \\
\midrule
\multicolumn{5}{l}{\textit{DS $-$ Base}}\\
~80\% & \textbf{$+$0.99} & \textbf{$+$3.51} & \textbf{$+$1.43} & $-$0.62 \\
~85\% & \textbf{$+$7.08} & \textbf{$+$1.68} & \textbf{$+$3.46} & \textbf{$+$1.32} \\
~90\% & \textbf{$+$9.54} & \textbf{$+$4.54} & \textbf{$+$3.53} & \textbf{$+$6.06} \\
\bottomrule
\end{tabular}
\end{table}

\subsection{Which Signals the LLM Kept}
\label{sec:whatlearned}
Because the reward is emitted as code, the selected function records which of the 14 evidence
signals offered on blocked rows (\S\ref{sec:dualstream}) the designer actually chose: it
references four, and relies on two.

The clearest case is its treatment of the legacy scores. The reward normalizes them onto the
newest models' scale and then never uses them again:

\begin{lstlisting}[language=Python, basicstyle=\ttfamily\scriptsize, backgroundcolor=\color{highlight}, numbers=none, frame=none]
# Legacy normalised (1-1000 -> 0-1), retained as
# secondary evidence only: deliberately NOT mixed
# into conf_score below.
legacy_g1_h1_n = legacy_g1_h1 / 1000.0
legacy_g2_h1_n = legacy_g2_h1 / 1000.0

# Primary confidence = newest (calibrated) head.
conf_score = torch.clamp(newest_h1, 0.0, 1.0)
\end{lstlisting}

\noindent Its docstring justifies the omission as avoiding ``stale-model noise''. The
newest-versus-legacy \emph{disagreement} schema we sketched in \S\ref{sec:dualstream} is therefore
not what the loop converged to: handed both, it promoted the newest calibrated head to sole
confidence term and demoted the legacy generations to validators---a staleness-aware choice no
fixed formula would make. Payment instrument and the three secondary heads of each generation are
never referenced. Buyer segment, which we offered without prescribing a use, is kept: the reward
conditions its block thresholds ($0.70$ for low-risk cohorts, $0.45$ for the highest-risk one) and
its reward weights on it, but never the confidence itself. We draw no conclusion that
the discarded signals are uninformative---only that exposing a superset and letting the designer
select is a workable way to find out, and that the choice stays auditable because it is code.

Two further behaviors emerged unprompted. Rather than scoring a linear
$(\text{score}-\text{threshold})$ margin, the reward \textbf{hinge-gates}---paying for blocking only
\emph{above} threshold and for passing only \emph{below}, with symmetric penalties in the crossover
band (intermediate terms inlined; exact code in Appendix~\ref{appendix:reward}):

\begin{lstlisting}[language=Python, basicstyle=\ttfamily\scriptsize, backgroundcolor=\color{highlight}, numbers=none, frame=none]
# BLOCK: hinge reward above threshold; penalty below.
block_pen_thr = seg_threshold - pass_margin - 0.15
r_block = torch.clamp(conf_score - seg_threshold, min=0.0) \
        - torch.clamp(block_pen_thr - conf_score, min=0.0) * 2.0

# PASS: the mirror image.
r_pass = torch.clamp(seg_threshold - conf_score, min=0.0) \
       - torch.clamp(conf_score - (seg_threshold + pass_margin),
                     min=0.0) * 2.0
\end{lstlisting}

\noindent Its docstring records this as a fix for the false positives a linear margin produced.
Finally it applies a \textbf{strong step-0 bias}, scaling blocked-stream rewards at
pre-authorization and adding a flat early-block bonus:

\begin{lstlisting}[language=Python, basicstyle=\ttfamily\scriptsize, backgroundcolor=\color{highlight}, numbers=none, frame=none]
# Strong step-0 bias to drive early (pre-auth) blocking.
if current_step == 0:
    step_mult, block_boost = 1.8, is_block * 0.5
else:
    step_mult, block_boost = 1.0, 0.0
r_blocked = seg_weight * (r_base * step_mult + block_boost)
\end{lstlisting}

\noindent This produces the pre-authorization interception shares of
Table~\ref{tab:oot_overall}, and echoes the early-stage up-weighting independently re-derived by
\citet{qu2025llmenhanced}. The labeled stream converged to a precision-leaning Lagrangian
($+2/{-}3/{-}2$ on TP/FP/FN), clamped to $[-2,2]$.

\section{Conclusion}
\citet{qu2025llmenhanced} showed that an LLM can evolve the reward of a two-step REINFORCE
fraud-blocking policy from labeled, approved traffic alone. We extend that framework to the
censored half of the problem: a single LLM evolves a \emph{dual-stream} reward spanning both
labeled approved and legacy-blocked transactions, performing reject inference inside the reward
itself and deciding for itself which risk signals to trust. Across a four-month OOT window the
resulting policy beats the production GBM baseline and the labeled-only agent at every target
recall in the operating range, by $+6.83$ and $+3.36$ points of dollar-weighted precision on
average, while leaving the policy architecture of \citet{qu2025llmenhanced}, and hence its reported
sub-50\,ms CPU inference, unchanged. The censored population carries usable risk signal; an LLM
can write the reward that extracts it, and leave a readable record of what it used.

\section*{Limitations}
\paragraph{Reject-inference bias.} The blocked population is a biased, non-random (MNAR)
sample. Stream~2's rewards assume the newest GBM scores carry genuine risk signal on blocked
rows; where the original block was driven by features the newest models do not observe, this
assumption can propagate legacy-policy bias. We caution against treating stream~2 rewards as
ground-truth labels.

\section*{Ethics Statement}
\paragraph{Data and privacy.} All experiments use internal historical transaction records
accessed under the platform's data-governance program; no human-subject data were
collected and no personally identifying attributes are used as model features.

\paragraph{Risk of harm.} Blocking errors carry real consequences: a false block harms a
legitimate buyer; a missed fraud harms sellers and the platform. Any deployment should include
human review, redress paths, and ongoing monitoring for differential impact across buyer
cohorts.

\bibliography{custom_v2}

\appendix
\section{Reward-Optimization Loop}
\label{appendix:alg}
Algorithm~\ref{alg:LLM-RL} details the LLM reward-optimization loop used throughout the main
text; the evolutionary outer loop is unchanged from that of \citet{qu2025llmenhanced}, and only the per-candidate reward (Stream~1$+$Stream~2) differs.

\begin{algorithm}[t]
\caption{LLM-based Dual-Stream Reward Optimization}
\label{alg:LLM-RL}
{\footnotesize
\begin{algorithmic}[1]
\REQUIRE $N_\text{iter}, N_\text{samples}, N_\text{episodes}, \theta_\text{recall}$
\STATE Initialize env.\ $\mathcal{E}$, baseline $\mathcal{M}_b$, best reward $f_\text{best}$, temperature
\FOR{$\text{iter}=1$ to $N_\text{iter}$}
  \STATE Update temperature from feedback
  \FOR{$k=1$ to $N_\text{samples}$}
    \STATE Sample $f^k$ from LLM; validate signature via AST; re-sample if invalid
  \ENDFOR
  \FORALL{valid $f^k$}
    \STATE $\mathcal{A}_k \leftarrow \textsc{Train}(\mathcal{E}, f^k, N_\text{episodes})$ \COMMENT{dual-stream data}
    \STATE $\text{feedback}_k \leftarrow \textsc{Eval}(\mathcal{A}_k, \mathcal{M}_b, \theta_\text{recall})$ on eval. set
  \ENDFOR
  \STATE Update $f_\text{best}$; update LLM instructions with best / sub-optimal / reflection
\ENDFOR
\end{algorithmic}}
\end{algorithm}

\section{Best Dual-Stream Reward Function}
\label{appendix:reward}
Listing~\ref{lst:stream2} gives the complete reward function selected by the loop
(iteration~25), reproduced verbatim apart from anonymization: the four GBM head names are
rendered \texttt{h1}--\texttt{h4}, the two legacy model generations \texttt{g1}/\texttt{g2},
and every buyer-segment descriptor is redacted to \texttt{\{descr\}}. The numeric coefficients
are the LLM's own evolved values and are unmodified. The three behaviors discussed in
\S\ref{sec:whatlearned} are visible here: the newest calibrated head is the sole confidence term
(the normalized legacy variables are computed but deliberately never mixed in), blocking and
passing are hinge-gated with symmetric crossover penalties, and step~0 carries a $1.8\times$
multiplier plus a fixed early-block bonus. The listing also retains the per-cohort threshold and
weight vectors the LLM derived from the injected fraud-rate statistics; these are peripheral to
the censored-outcome mechanism this paper is about, and are reproduced only so that the artifact
is shown complete and unedited.

\begin{lstlisting}[language=Python, basicstyle=\ttfamily\scriptsize, breaklines=true, backgroundcolor=\color{highlight}, numbers=none, caption={Best dual-stream reward (iteration~25), anonymized. \texttt{\{descr\}} marks a redacted buyer-segment descriptor.}, label={lst:stream2}]
import torch

def get_reward(current_step, action, target, wgt, is_blocked,
               legacy_g1_h1, legacy_g1_h2, legacy_g1_h3, legacy_g1_h4,
               legacy_g2_h1, legacy_g2_h2, legacy_g2_h3, legacy_g2_h4,
               newest_h1, newest_h2, newest_h3, newest_h4,
               byr_segment, pymnt_inst):
    """
    Precision-gated score thresholding for the blocked stream.

    Stream 1 (labeled): dollar-weighted Lagrangian, precision-leaning
        (TP +2, FP -3, FN -2), with a mild step-0 boost (1.05).
    Stream 2 (blocked, censored): the newest calibrated GBM head is the
        PRIMARY confidence; legacy scores are kept only as SECONDARY
        validators and deliberately NOT blended into the confidence, to
        avoid injecting stale-model noise.
    Per-segment block thresholds are set INVERSELY to the injected
        per-segment precision statistics; segment weights likewise.
    Hinge gating: reward blocking only above threshold and passing only
        below, with symmetric penalties in the crossover band.
    Strong step-0 bias on the blocked stream (step_mult 1.8, block_boost 0.5).
    Final clamp to [-2, 2] for REINFORCE gradient stability.
    """
    device = action.device
    reward = torch.zeros_like(action.float(), device=device)

    labeled_mask = (is_blocked == 0.0).float()
    blocked_mask = (is_blocked == 1.0).float()
    a, t = action.long(), target.long()

    # ---- STREAM 1: LABELED (ground truth available) ----
    if labeled_mask.sum() > 0:
        is_tp = ((a == 1) & (t == 1)).float()
        is_fp = ((a == 1) & (t == 0)).float()
        is_fn = ((a == 0) & (t == 1)).float()
        r_labeled = (is_tp * 2.0 + is_fp * -3.0 + is_fn * -2.0) * wgt
        step_boost = 1.05 if current_step == 0 else 1.0
        reward = torch.where(labeled_mask == 1.0, step_boost * r_labeled, reward)

    # ---- STREAM 2: BLOCKED (label unobservable; reason over scores) ----
    if blocked_mask.sum() > 0:
        # Legacy normalised (1-1000 -> 0-1), retained as secondary evidence
        # only: deliberately NOT mixed into conf_score below.
        legacy_g1_h1_n = legacy_g1_h1 / 1000.0
        legacy_g2_h1_n = legacy_g2_h1 / 1000.0

        # Primary confidence = newest (calibrated) head.
        conf_score = torch.clamp(newest_h1, 0.0, 1.0)
        seg_float = byr_segment.float()

        # Per-segment BLOCK thresholds, set inversely to segment precision.
        t_seg_0 = 0.70 * (seg_float == 0).float()   # {descr}  (strictest)
        t_seg_1 = 0.60 * (seg_float == 1).float()   # {descr}
        t_seg_2 = 0.50 * (seg_float == 2).float()   # {descr}
        t_seg_3 = 0.65 * (seg_float == 3).float()   # {descr}
        t_seg_4 = 0.45 * (seg_float == 4).float()   # {descr}  (most lenient)
        t_seg_5 = 0.80 * (seg_float == 5).float()   # {descr}
        seg_threshold = (t_seg_0 + t_seg_1 + t_seg_2
                         + t_seg_3 + t_seg_4 + t_seg_5)

        # Per-segment reward weights.
        w_seg_0 = 1.0 * (seg_float == 0).float()    # {descr}
        w_seg_1 = 1.0 * (seg_float == 1).float()    # {descr}
        w_seg_2 = 1.2 * (seg_float == 2).float()    # {descr}
        w_seg_3 = 0.8 * (seg_float == 3).float()    # {descr}
        w_seg_4 = 1.5 * (seg_float == 4).float()    # {descr}
        w_seg_5 = 0.5 * (seg_float == 5).float()    # {descr}
        seg_weight = (w_seg_0 + w_seg_1 + w_seg_2
                      + w_seg_3 + w_seg_4 + w_seg_5)

        pass_margin = 0.05
        is_block = (a == 1).float()
        is_pass = (a == 0).float()

        # BLOCK: hinge reward above threshold; penalty far below it.
        block_reward = torch.clamp(conf_score - seg_threshold, min=0.0)
        block_pen_thr = seg_threshold - pass_margin - 0.15
        block_penalty = torch.clamp(block_pen_thr - conf_score, min=0.0) * 2.0
        r_block = block_reward - block_penalty

        # PASS: hinge reward below threshold; penalty far above it.
        pass_reward = torch.clamp(seg_threshold - conf_score, min=0.0)
        pass_penalty = torch.clamp(
            conf_score - (seg_threshold + pass_margin), min=0.0) * 2.0
        r_pass = pass_reward - pass_penalty

        r_base = is_block * r_block + is_pass * r_pass

        # Strong step-0 bias to drive early (pre-auth) blocking.
        if current_step == 0:
            step_mult, block_boost = 1.8, is_block * 0.5
        else:
            step_mult, block_boost = 1.0, 0.0

        r_blocked = seg_weight * (r_base * step_mult + block_boost)
        reward = torch.where(blocked_mask == 1.0, r_blocked, reward)

    return torch.clamp(reward, min=-2.0, max=2.0)
\end{lstlisting}

\section{Prompt Design}
\label{appendix:prompts}
This appendix reproduces the natural-language templates that drive the LLM
reward-design loop (Figure~\ref{fig:LLM-RL-overview}). Two runtime placeholders
are filled programmatically before each query:
{\small\texttt{\{task\_reward\_signature\_string\_baseline\}}} is replaced by the
validated 4-argument SS reward (the Stream~1 starting point), and
\texttt{\{task\_reward\_signature\_string\}} by the running best-so-far
19-argument dual-stream reward. Proprietary specifics are anonymized to match the main
body: the concrete buyer-cohort and payment-instrument \emph{category names} are
replaced by generic lifecycle / instrument classes, and the absolute per-cohort
dollar-weighted fraud rates injected as ``statistics context'' are replaced by
normalized relative-risk indices spanning roughly an order of magnitude across
cohorts (absolute rates withheld). The integer \texttt{byr\_segment} codes below
are the raw model enumeration, an arbitrary and \emph{non} risk-sorted ordering. The
four legacy risk-model heads are denoted $h_1$--$h_4$, following the same
anonymization used by \citet{qu2025llmenhanced} (their internal fraud-typology
names are withheld).
All other text is verbatim, with non-ASCII math glyphs rendered in ASCII
($\in\rightarrow$\texttt{in}, $\times\rightarrow$\texttt{x},
$\rightarrow\rightarrow$\texttt{->}, $\approx\rightarrow$\texttt{\textasciitilde}).

\subsection{Dual-Stream Task Description}
\label{appendix:prompt-task}
The dual-stream task prompt frames the two-population problem and enumerates
the reward signals available on blocked, censored rows.

\begin{lstlisting}[basicstyle=\ttfamily\scriptsize,breaklines=true,numbers=none,frame=single,captionpos=b,caption={Dual-stream task description prompt (anonymized, ASCII).},label={lst:prompt_task}]
Design a reward function that enables the RL agent to make more effective
two-step fraud-blocking decisions for improved performance on the held-out
evaluation population, compared to the baseline GBM score at step 0.

The training data contains TWO types of samples:

(1) LABELED samples (is_blocked == 0.0):
    Transactions approved by the legacy system. Actual fraud label is known
    (target in {0, 1}). Apply standard Lagrangian precision/recall reward
    logic, identical to the labeled-only baseline. Legacy scores are 0.0 for these
    rows -- ignore them. byr_segment and pymnt_inst are 0 here -- ignore.

(2) BLOCKED samples (is_blocked == 1.0):
    Transactions blocked by the legacy system. Actual fraud label is
    UNOBSERVABLE (target == -1). Use these signals to estimate the reward
    for the agent's action:

    Legacy model scores (1-1000 scale; PRA at step 0, PA at step 1):
        legacy_g1_{h1, h2, h3, h4}  -- legacy gen-1 GBM scores
        legacy_g2_{h1, h2, h3, h4}  -- legacy gen-2 GBM scores
    Newest GBM scores (0-1 scale, step-appropriate):
        newest_{h1, h2, h3, h4}     -- current production GBM scores

    SCALE NOTE: legacy scores are 1-1000, newest are 0-1. Normalise before
    comparison: legacy_g1_h1 / 1000.0 ~ newest_h1 range. Example: legacy_g1_h1
    > 700 is equivalent to newest_h1 > 0.7.

    HOW TO INTERPRET SCORE SIGNALS:
      - Both legacy and newest HIGH -> strong fraud evidence  -> reward blocking
      - Legacy HIGH, newest LOW     -> possible over-block     -> discount reward
      - Both scores LOW             -> likely legacy false pos -> reward passing
      - gen-1 and gen-2 agreeing HIGH adds more confidence than either alone

    Context tensors (blocked stream only; 0 for labeled rows):
      byr_segment : int64 buyer-segment code (label-anonymized generic
                    lifecycle classes; arbitrary, non risk-sorted enumeration):
          0 = established, regularly active buyer
          1 = occasional buyer
          2 = long-dormant before recent purchase (historically over-blocked)
          3 = new account, limited history (high legacy block rate)
          4 = guest checkout (no account login)
          5 = unknown
      pymnt_inst  : int64 payment-instrument code (instrument class only;
                    brands and regions withheld):
          0 = credit card        1 = debit card
          2 = mobile wallet A     3 = mobile wallet B
          4 = regional real-time bank transfer
          5 = direct bank-account debit
      Dollar-weighted fraud rates per segment, payment method, and their
      combination are provided in the injected statistics context (the
      normalized relative-risk form; absolute rates withheld).

GOAL:
    Drive the RL agent to block more high-value fraud at step 0 (pre-auth)
    than step 1 (post-auth), and improve dollar-weighted precision at target
    recall on the evaluation population. The blocked-stream reward should not
    destabilise labeled-stream learning -- keep stream-2 reward magnitudes
    comparable to stream-1 (use wgt weighting for both).
\end{lstlisting}

\subsection{Code-Output Instruction (19-Argument Contract)}
\label{appendix:prompt-codetip}
The code-output prompt pins the exact function signature, the TorchScript
constraint, and the extraction format.

\begin{lstlisting}[basicstyle=\ttfamily\scriptsize,breaklines=true,numbers=none,frame=single,captionpos=b,caption={Code-output instruction prompt (anonymized, ASCII). The two backtick-fenced placeholders are filled at runtime.},label={lst:prompt_codetip}]
Your reward function should use useful variables from the code framework as
inputs. As a reference, below are validated reward functions designed for
labeled-data transactions only (Stream 1, is_blocked==0.0). They use only 4
arguments (current_step, action, target, wgt) and do NOT handle blocked
samples -- use them as a starting point for the Stream 1 logic only, then
design the Stream 2 reward (blocked samples, is_blocked==1.0) yourself, using
the legacy scores, newest GBM scores, byr_segment, and pymnt_inst:
```python {task_reward_signature_string_baseline} ```
The best full dual-stream reward function (both streams) found so far is:
```python {task_reward_signature_string} ```
Since the reward function will be compiled with TorchScript, ensure the code
is TorchScript-compatible (use torch tensors, not numpy). Ensure any new
tensor or variable is on the same device as the inputs. The output must be:
    (1) the completed reward function.
    (2) the reward code must use this exact 19-argument signature:
        "def get_reward(current_step, action, target, wgt, is_blocked,
         legacy_g1_h1, legacy_g1_h2, legacy_g1_h3, legacy_g1_h4,
         legacy_g2_h1, legacy_g2_h2, legacy_g2_h3, legacy_g2_h4,
         newest_h1, newest_h2, newest_h3, newest_h4,
         byr_segment, pymnt_inst):"
        -- all 19 arguments are required and must match exactly.
    (3) the code output must be formatted as "```python ... ```".
    (4) if you define extra helper functions, output them completely in the
        same code block.
    (5) code and annotations must be consistent.
    (6) output only the reward-function code at the beginning of the response,
        for ease of extraction.
    (7) use the backslash (\) as a line continuation where one logic line is
        split across physical lines.
    (8) always add "import torch" as the first line of the code block.
\end{lstlisting}

\subsection{Reward-Signature Argument Reference}
\label{appendix:prompt-sig}
The reward-signature reference is the per-argument documentation the model
conditions on; a full example reward is given as Listing~\ref{lst:stream2}
(Appendix~\ref{appendix:reward}).

\begin{lstlisting}[basicstyle=\ttfamily\scriptsize,breaklines=true,numbers=none,frame=single,captionpos=b,caption={Condensed 19-argument reward-signature reference (anonymized, ASCII).},label={lst:prompt_sig}]
current_step : int -- 0 = pre-auth (PRA, all txns); 1 = post-auth (PA, only
                      txns passed at step 0)
action  : Tensor (N,) int64 -- 0=pass, 1=block
target  : Tensor (N,) int64 -- 0=legit, 1=fraud, -1=blocked (unknown)
wgt     : Tensor (N,) float -- dollar weight = usd_amt x sample_weight
is_blocked : Tensor (N,) float -- 1.0 = legacy-blocked (no ground truth);
                                  0.0 = labeled approved

Legacy scores (1-1000 scale, original production scale, NOT normalised), one
per GBM head {h1, h2, h3, h4} for each legacy generation {g1, g2}:
   legacy_g1_h1 .. legacy_g1_h4 , legacy_g2_h1 .. legacy_g2_h4
   -- score at the current step's checkpoint (PRA at step 0, PA at step 1).
      For labeled rows (is_blocked==0) these are 0.0. Legacy scores are NOISY
      evidence: the block decision followed complex threshold rules that these
      scores only partially encode.

Newest GBM scores (0-1 normalised; also used as policy input):
   newest_h1, newest_h2, newest_h3, newest_h4  -- current-step scores.

Context (blocked stream only; 0 for labeled rows):
   byr_segment : int64 buyer-segment code (generic lifecycle classes; arbitrary,
       non risk-sorted enumeration):
       0=established-active  1=occasional  2=long-dormant (historically
       over-blocked -- use conservatively)  3=new-account (high legacy block
       rate -- watch false positives)  4=guest-checkout  5=unknown
   pymnt_inst  : int64 payment-instrument code (class only; brands/regions
       withheld): 0=credit card  1=debit card  2=mobile wallet A
       3=mobile wallet B  4=regional bank transfer (mostly low-risk)
       5=direct bank-account debit

Score-scale bridging (legacy 1-1000 vs newest 0-1): normalise legacy, e.g.
   legacy_g1_h1_norm = legacy_g1_h1 / 1000.0 ; threshold 700/1000 = 0.7.
Useful derived quantities (blocked stream):
   score_agreement_h1 = 1.0 - abs(newest_h1 - legacy_g1_h1_norm)
       # 1.0 = agree, 0.0 = maximally disagree
   risk_level = newest_h1*0.5 + (legacy_g1_h1_norm + legacy_g2_h1_norm)*0.25
       # newer GBM weighted higher; clamp to [0,1]

Design principles:
   STREAM 1 (is_blocked==0.0): standard Lagrangian precision/recall reward,
       identical to the labeled-only baseline; target in {0,1}; ignore legacy scores and context.
   STREAM 2 (is_blocked==1.0): target==-1 (unobservable). Reward from
       (a) legacy/newest agreement  -> strong fraud signal;
       (b) disagreement (legacy high, newest low) -> possible over-block -> discount;
       (c) both low -> likely legacy false positive -> reward passing;
       (d) condition on byr_segment and pymnt_inst.
TorchScript: use torch.* (not numpy); cast integer comparisons to float,
   e.g. (byr_segment == 2).float(); torch.clamp / abs / where are safe.
Return: reward  # Tensor (N,) or (N,1), aggregated externally.
\end{lstlisting}

\section{Overall Precision--Recall Curve}
\label{appendix:oot}
Table~\ref{tab:app_curve} gives the full swept operating curve for both LLM-designed agents on the
held-out April--July~2026 window, extending beyond the three targets of
Table~\ref{tab:oot_overall}. Across the rows spanning recall $0.65$--$0.90$, which contains the
reported band, DS attains higher \$precision at comparable \$recall; outside that span---below
$0.63$ and above $0.95$, where neither agent is intended to operate---the ordering is mixed.

\begin{table}[!ht]
\centering\small
\setlength{\tabcolsep}{7pt}
\caption{Full swept curve on the held-out April--July~2026 window for \textbf{SS} (labeled-only)
and \textbf{DS} (dual-stream). Each row is one aligned operating point, ordered by increasing
\$Recall; the two agents' achieved recalls differ slightly, so each agent's pair should be read
together rather than compared column-wise. Baseline values at the three reported targets are in
Table~\ref{tab:oot_overall}.}
\label{tab:app_curve}
\begin{tabular}{rr@{\hskip 12pt}rr}
\toprule
\multicolumn{2}{c}{SS} & \multicolumn{2}{c}{\textbf{DS}} \\
\cmidrule(lr){1-2}\cmidrule(lr){3-4}
\$Recall & \$Prec & \$Recall & \$Prec \\
\midrule
0.5773 & 0.6788 & 0.5012 & 0.6690 \\
0.5773 & 0.6788 & 0.5646 & 0.6629 \\
0.6256 & 0.6565 & 0.6117 & 0.6479 \\
0.6513 & 0.6120 & 0.6594 & 0.6394 \\
0.7016 & 0.5886 & 0.7107 & 0.6183 \\
0.7513 & 0.5581 & 0.7548 & 0.5825 \\
0.8001 & 0.5041 & 0.8007 & 0.5515 \\
0.8530 & 0.4400 & 0.8510 & 0.4636 \\
0.9027 & 0.3295 & 0.9011 & 0.3592 \\
0.9553 & 0.1581 & 0.9517 & 0.1349 \\
\bottomrule
\end{tabular}
\end{table}

\end{document}